\documentclass{bmvc2k}
\usepackage{bbm}
\usepackage{multirow}
\usepackage{booktabs}
\usepackage{amssymb}
\usepackage{dsfont}
\usepackage{indentfirst} 
\usepackage{hyperref}
\usepackage[capitalize]{cleveref}
\usepackage{titlesec}
\usepackage{makecell}
\usepackage{pifont}
\titleformat*{\section}{\large\bfseries}
\titleformat*{\subsection}{\normalsize\bfseries}
\newcommand{\cmark}{\ding{51}}%
\newcommand{\xmark}{\ding{55}}%
\newcommand{\OURS}{XCon}

\title{\OURS: Learning with Experts for Fine-grained Category Discovery}
\addauthor{Yixin Fei}{yixin.feiyx@gmail.com}{1}
\addauthor{Zhongkai Zhao}{zhongkai.zhaok@gmail.com}{1}
\addauthor{Siwei Yang}{swyang.ac@gmail.com}{1,3}
\addauthor{Bingchen Zhao}{zhaobc.gm@gmail.com}{2,3}

\addinstitution{
 Tongji University\\
 Shanghai, China
}
\addinstitution{
 University of Edinburgh,\\
 Edinburgh, UK
}
\addinstitution{
 LunarAI
}

\runninghead{Y. Fei \etal}{Learning with Experts for Fine-grained Category Discovery}

\def\eg{\emph{e.g}\bmvaOneDot}

\def\etal{\emph{et al}\bmvaOneDot}
\def\ie{\emph{i.e}\bmvaOneDot}

\titlespacing*{\section}{0.2pt}{*1}{*0}
\titlespacing*{\subsection}{0pt}{*1}{*0}
\begin{document}

\maketitle

\begin{abstract}
We address the problem of generalized category discovery (GCD) in this paper, \ie clustering the unlabeled images leveraging the information from a set of seen classes, where the unlabeled images could contain both seen classes and unseen classes.
The seen classes can be seen as an implicit criterion of classes, which makes this setting different from unsupervised clustering where the cluster criteria may be ambiguous.
We mainly concern the problem of discovering categories within a fine-grained dataset since it is one of the most direct applications of category discovery, \ie helping experts discover novel concepts within an unlabeled dataset using the implicit criterion set forth by the seen classes.
State-of-the-art methods for generalized category discovery leverage contrastive learning to learn the representations, but the large inter-class similarity and intra-class variance pose a challenge for the methods because the negative examples may contain irrelevant cues for recognizing a category so the algorithms may converge to a local-minima.
We present a novel method called Expert-Contrastive Learning (\OURS) to help the model to mine useful information from the images by first partitioning the dataset into sub-datasets using $k$-means clustering and then performing contrastive learning on each of the sub-datasets to learn fine-grained discriminative features.
% TODO: refine above
Experiments on fine-grained datasets show a clear improved performance over the previous best methods, indicating the effectiveness of our method.
\end{abstract}

%-------------------------------------------------------------------------
\section{Introduction}
\label{sec:intro}
% intro
Deep learning models have achieved super-human performance on many computer vision problems where large-scale human annotations are available, such as image recognition~\cite{deng2009imagenet} and object detection~\cite{ren2015faster}.
However, collecting a dataset at scales like ImageNet or COCO is not always possible.
Consider the scenario of fine-grained recognition such as bird species recognition or medical image analysis, where the annotations require expert knowledge which could be costly to collect, also it is difficult for the collected annotations to cover all the possible classes because new classes keep growing over time.

% setting, difference with ssl
The problem of generalized category discovery was recently formalized in~\cite{vaze2022generalized}, where the aim is to discover categories within the unlabeled data by leveraging the information from a set of labeled data.
It is assumed that the labeled data contains similar yet distinct classes from the unlabeled data.
The labeled data collected by human experts can be seen as an implicit criterion of classes which can be learned by the model to perform clustering on the unlabeled data.
This setting is much harder than semi-supervised learning because generalized category discovery does not assume we know all the classes in the data while in semi-supervised learning the assumption is that the labeled data covers all the classes including ones in unlabeled data.

\begin{figure*}[t]
\begin{center}
\includegraphics[width=\textwidth]{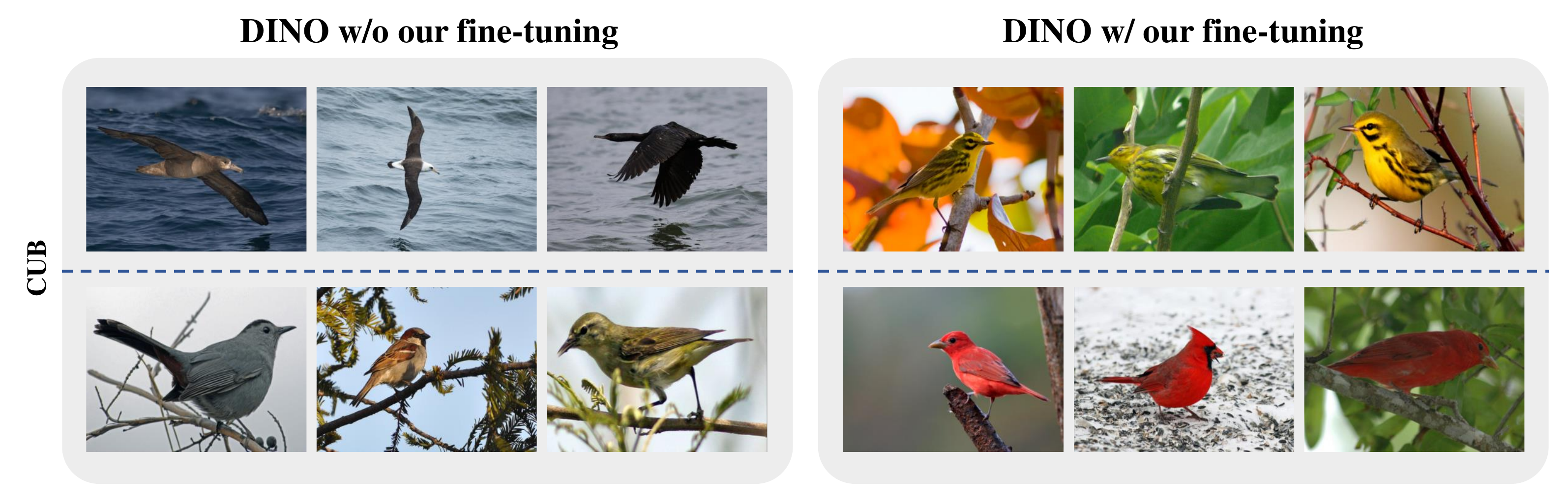}
\end{center}
\vspace{-22pt}
\caption{$k$-means results on the DINO features and features after fine-tuning with our method. The images in each row represent a cluster in $k$-means. The clusters formed by DINO features are mainly based on the class irrelevant cues, \eg, background and object pose. The features learned by our method could cluster the images based on the correct cues, the object classes.}
\label{fig:kmeans}
\end{figure*}

% intro to fine-grain classes and challenges
In this paper, we specifically focus on fine-grained generalized category discovery which is a more difficult and practical problem than generic category discovery since field experts are interested in the fine-grained concepts in real applications, and they often have a labeled dataset representing the existing knowledge, so such a fine-grained generalized category discovery method could help them make sense of the unlabeled set by clustering the unlabeled instance according to the criteria implicitly defined in the labeled data.
In fine-grained category discovery, the main challenge is the large inter-class similarity and the intra-class variance, different classes may require the model to learn more discriminative features to be able to distinguish, \eg, two different birds could only differ in the beak.
We have observed that an unsupervised representation (\eg DINO) could cluster the data based on class irrelevant cues such as the object pose or the background, see the left part of~\cref{fig:kmeans}.
Based on this observation, we proposed a simple yet effective method to boost the performance of generalized category discovery on fine-grained data named Expert Contrastive Learning (\textbf{\OURS}).

% our contribution
In our proposed \OURS~method, we partition the data into $k$ expert sub-datasets by directly performing $k$-means clustering on self-supervised representations. These $k$ sub-datasets can be used as a strong prior for the next learning phase because within each of the $k$ sub-datasets, class-irrelevant cues will be so similar that the model will be forced to learn more class-relevant features within each sub-dataset.
Each of these sub-datasets can be viewed as an expert dataset used to eliminate the negative influence introduced by certain kinds of class-irrelevant cues.
To learn a robust representation from these datasets, we directly leverage supervised contrastive learning~\cite{khosla2020supervised} on the labeled data and unsupervised contrastive learning~\cite{chen2020simple} on all the data.

Our contribution is three-fold:
\begin{itemize}
    \item We observed that self-supervised representations can group the data based on class irrelevant cues which can be exploited to design further methods.
    \item We proposed a method that can learn discriminative features for fine-grained category discovery by partitioning the data into $k$ sub-datasets.
    \item We validated the effectiveness of our proposed method by setting a new state-of-the-art performance on seven tested generalized category discovery benchmarks.
\end{itemize}

Our code is available at \url{https://github.com/YiXXin/XCon}.

\section{Related Works}
\label{sec:related}

\subsection{Novel Category Discovery}
%Novel Category Discovery(NCD)  aims at training a model to discover new classes in the unlabeled samples based on the knowledge of labeled samples. 
Novel Category Discovery (NCD) aims to discover new object categories by transferring the knowledge learned from a set of relevant but different seen classes.
This task was first formalized in DTC~\cite{han2019learning}, with earlier works~\cite{hsu2017learning,hsu2019multi} tackling a similar problem.
KCL~\cite{hsu2017learning} and MCL~\cite{hsu2019multi} utilize the pairwise similarity to transfer the clustering model to cross-task scenarios, which can be used to categorize the unseen classes further.
%With the realistic setting, the typical two-step method introduced in DTC~\cite{han2019learning} is the representation learning on the labeled data and the fine-tuning on the unlabeled data by using the deep embedded clusterings.
A common three-step learning pipeline is proposed in RankStat~\cite{Han2020Automatically} where the representation is first learned with self-supervision on all the data and then fine-tuned on the labeled data, the final representation used for discovering novel categories is then further fine-tuned using a pair-wise clustering loss on the unlabeled data.
Since then, many works~\cite{zhao2021novel, Han2020Automatically, zhong2021neighborhood, chi2021meta, fini2021unified, zhong2021openmix} begin to focus on this NCD problem and present promising results.
% some work leverages three step learning pipeline~\cite, contrastive ~\cite, two-branch~\cite, meta-learning~\cite
%Han \etal~\cite{Han2020Automatically} further combined a three step learning pipeline.
%They first train labeled and unlabeled data to learn an unbiased image representation, then only fine-tune on the labeled data to learn higher level features and finally use rank statistics for robust clustering.
Contrastive learning has been explored under this NCD problem by NCL~\cite{zhong2021neighborhood}, showing strong performance.

Efforts have also been made in extending this problem to the more challenging fine-grained classification scenario by DualRank~\cite{zhao2021novel}, which leverages the local object parts information to enhance the representations used for discovering novel categories.
%NCL~\cite{zhong2021neighborhood} adopts contrastive learning to obtain a stronger information of positive samples and generates hard negatives through feature mixing.
%DualRank~\cite{zhao2021novel} propose a two-branch model that focus on local information and overall feature separately leveraging dual ranking statistic and mutual knowledge distillation.
%Meta Discovery~\cite{chi2021meta} find that novel classes share high-level semantic features with known calsses and then link NCD to meta-learning.
%There are also some methods~\cite{fini2021unified, zhong2021openmix} that treat the problem as a unified framework.
%A multi-view self-labeling strategy is implemented in UNO~\cite{fini2021unified} and a combination of labeled and unlabeled data as well as their labels and pseudo-labels is proposed in Openmix~\cite{zhong2021openmix}. 
Our work also focuses on the challenging fine-grained classification scenario. The key difference with prior works is that we use $k$-means grouping on a self-supervised feature to provide informative pairs for contrastive learning instead of using MixUp~\cite{zhong2021neighborhood} or local object parts~\cite{zhao2021novel}.
Our work also builds on a newly proposed setting named Generalized Category Discovery~(GCD)~\cite{vaze2022generalized} where the unlabeled examples can come from both seen and unseen classes, which is a more realistic scenario than NCD.
%GCD
%Recently, a more unconstrained and realistic setting is considered that the unlabeled samples may come from both known and unknown categories, instead of only unknown categories, termed as Generalized Category Discovery(GCD)~\cite{vaze2022generalized}.
%Vaze \etal propose a baseline of using vision transformers with contrastive representation learning.
%ComEX ~\cite{yang2022divide} uses two groups of experts to divide and conquer the problem.

\subsection{Contrastive Learning}
% MoCo, SimCLR, DINO (use this for initial features)
% SupCon, MoCoFT.
Contrastive learning has been showing to be effective for learning representations~\cite{he2019momentum,chen2020simple} in a self-supervised manner using the instance discrimination pretext~\cite{wu2018unsupervised} as the learning objective.
Instance discrimination learns the representation by pushing negative examples away from each other and pulling positive examples closer in the embedding space.
As informative examples are important for learning representations with contrastive learning, there are works following this direction trying to create more informative negative or positive pairs using MixUp~\cite{zhu2021improving,kalantidis2020hard} or special augmentations~\cite{shen2022unmix}.

Our focus is to learn representations that can be used to discover novel fine-grained categories within the unlabeled dataset, for which a strong representation is needed.
By creating informative contrastive pairs by partitioning the dataset into $k$ sub-datasets using $k$-means, examples within each sub-dataset will be similar so that the model will be forced to learn more discriminative features.
Compared to previous GCD methods with contrastive learning~\cite{vaze2022generalized}, our method shows clear performance improvements.

%\subsection{Semi-Supervised Learning}

%\subsection{Mixture-of-Experts}
% metric learning divide and conquer
% RIDE
% other works.

\section{Methods}

\begin{figure*}[htb]
\begin{center}
\includegraphics[width=\textwidth]{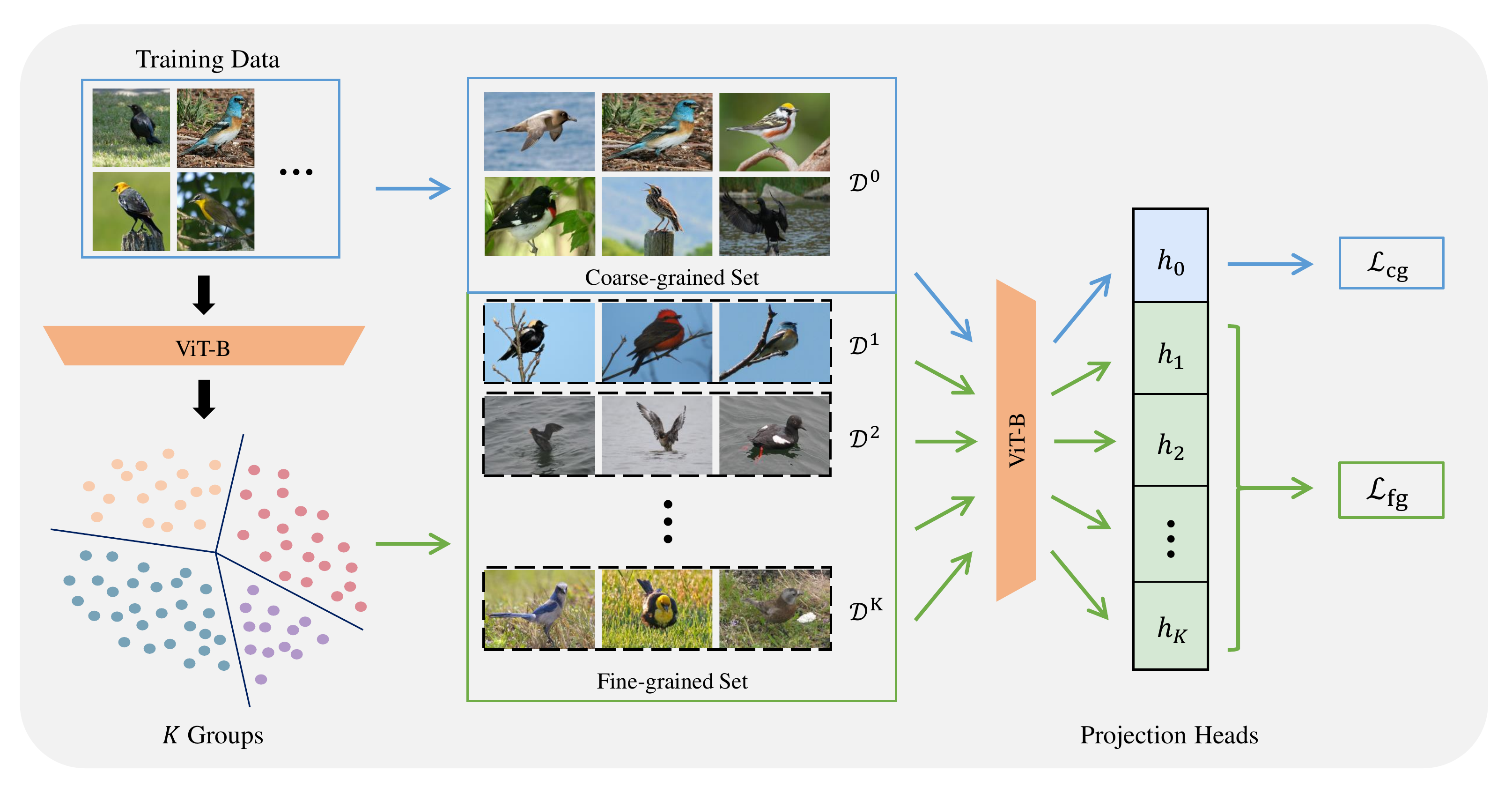}
\end{center}
\vspace{-22pt}
\caption{Overview of our \OURS~framework. We first partition the dataset into $K$ sub-datasets using $k$-means clustering the DINO~\cite{caron2021emerging} pretrained representations, then we perform joint contrastive representation learning on each of the partitioned sub-datasets $\mathcal{D}^{1}\dots \mathcal{D}^{K}$ as well as on the full dataset $\mathcal{D}^0$. Each of the partitioned sub-datasets will force the model to learn fine-grained discriminative information, because the background is similar within each of the sub-datasets so the model will need to learn the difference on the objects to be able to distinguish the examples.}
\label{fig:overview}
\end{figure*}
% problem definition
In GCD, the training dataset contains two parts, a labeled dataset $\mathcal{D}^l=\left\{(\mathbf{x}_i^l, y_i^l)\right\}$ and an unlabeled dataset $\mathcal{D}^u=\left\{(\mathbf{x}_i^u, y_i^u)\right\} $, where $y_i^l\in \mathcal{C}^l$ and $y_i^u\in \mathcal{C}^u$.
$\mathcal{C}^l$ are only composed of seen classes while $\mathcal{C}^u$ are composed of both seen and unseen classes, thus $\mathcal{C}^l \subseteq \mathcal{C}^u$.  
The goal of GCD is to learn a model to categorize the instances in $\mathcal{D}^u$ by leveraging the information from $\mathcal{D}^l$.
Compared to the previous NCD problem that considers the class sets as $\mathcal{C}^l \cap \mathcal{C}^u=\emptyset$, GCD is more challenging and practical.
%In addition, our real-world objects not only have the general information that categorize them in coarse-grained classes, but also possess some special features partialy that discriminate them from super-classes into sub-classes.
%The fine-grained classification is more significant and useful given the applications of GCD.

% Our focus, fine-grained GCD
It has been shown that self-supervised ViT features~\cite{caron2021emerging} could be a good initialization for representation learning in GCD~\cite{vaze2022generalized}.
In Vaze~\etal~\cite{vaze2022generalized}, contrastive learning is used to fine-tune the representation using the information from both labeled and unlabeled datasets, and it is shown that contrastive learning could indeed improve the performance of the representation on the task of GCD.
Informative contrastive pairs are important for representation learning, especially in the fine-grained classification setting where the model needs to learn subtle discriminative cues between categories.
We proposed a simple method that partitions the dataset into $k$ sub-datasets by using $k$-means on self-supervised features to help construct informative contrastive pairs for representations, the overview of our framework is shown in~\cref{fig:overview}.

\subsection{Preliminary}
In this section, we briefly review the method proposed in Vaze~\etal~\cite{vaze2022generalized} for GCD, which consists of two parts, representation learning and class assignment.
For representation learning, Vaze~\etal~\cite{vaze2022generalized} fine-tunes the representation by performing supervised contrastive learning on the labeled data and unsupervised contrastive learning on all the data to avoid outfitting the seen classes.%with a feature backbone $\textit{f}$ and a projection head $h$.
%In a minibatch $B$, there are two views $x_i$ and $x_{i}'$ of one same image.

The unsupervised contrastive loss is defined as
\begin{equation}\label{eq:unsup_con}
\mathcal{L}^{u}_{i} =
- \log \frac{\exp \left(\mathbf{z}_i \cdot \hat{\mathbf{z}}_{i}/ \tau\right)}{\sum_{n}  \mathds{1}_{[n \ne i]} \exp \left(\mathbf{z}_i \cdot \mathbf{z}_n / \tau\right)}
\end{equation}
where $\mathbf{z}_i=h(\textit{f}(x_i))$ is the feature extracted by a backbone $\textit{f}(\cdot)$ on the input image $x_i$ and projected to the embedding space via a projection head $h(\cdot)$, $\hat{\mathbf{z}}_i$ is the feature from another view of the input image $\hat{x}_i$.

The supervised contrastive loss is defined as
\begin{equation}\label{eq:sup_con}
\mathcal{L}^{s}_{i} =
- \frac{1}{|\mathcal{N}(i)|}  \sum_{q \in \mathcal{N}(i)}\log \frac{\exp \left(\mathbf{z}_i \cdot \mathbf{z}_q / \tau\right)}{\sum_{n} \mathds{1}_{[n \ne i]} \exp \left(\mathbf{z}_i \cdot \mathbf{z}_n / \tau\right)}
\end{equation}
where $\mathcal{N}(i)$ is the set of indices of images in the minibatch that have the same label $y_i$ with the anchor image $i$.

The final learning objective is the combination of these two losses
\begin{equation}
\label{eq:total}
\mathcal{L}_{\text{coarse}} = (1 - \lambda) \sum_{i \in \mathcal{B}_\mathcal{U} \cup \mathcal{B}_\mathcal{L}} \mathcal{L}^{u}_{i} +
\lambda \sum_{i \in \mathcal{B}_\mathcal{L}} \mathcal{L}^{s}_{i}
\end{equation}
where $\lambda$ is used to balance between these two terms, $\mathcal{B}_\mathcal{U}$ is a minibatch of unlabeled images, and $\mathcal{B}_\mathcal{L}$ is a minibatch of labeled images.

For class assignments, the semi-supervised $k$-means method is proposed.
The overall procedure is similar to the original $k$-means method~\cite{MackQueen67_Kmeans}, with a key difference that semi-supervised $k$-means is aware of the labeled data in $\mathcal{D}^l$, and in each step to recompute the cluster assignment, the samples that already have labels will be assigned to the correct cluster regardless of its distance to the nearest cluster centroids.
%The samples in $\mathcal{D}^l$ are assigned to the ground-truth centroids everytime the same, while other samples in $\mathcal{D}^u$ are assigned based on the distance.

\subsection{Dataset Partitioning}
The key challenge in representation learning for fine-grained GCD is that the representation is required to be sensitive to the detailed discriminative traits of different classes.
Learning the model by contrasting between examples in the full dataset may not help the model to learn such a discriminative representation.
Thus, we take advantage of the self-supervised representations that can roughly cluster the images according to the overall image statistics (\eg background, object pose, etc.)~\cite{caron2021emerging} to perform a preprocess on the full dataset by partitioning it into $k$ expert sub-datasets.
The overall statistics within each sub-dataset will be similar and then the model will naturally learn fine-grained discriminative features to distinguish between different examples within each sub-dataset.
Each of these expert sub-datasets will be expected to reduce different class-irrelevant cues represented by different overall image statistics.
%The goal of our dataset splitting is to cluster the features that are similar in coarse-grained classes.
%We can further learn about the discriminative fine-grained features within the sub-groups that have closely relevant coarse-grained features and are totally irrelevant to fine-grained classes.
%We introduce a way to pre-proccess the dataset for grouping the similar features together by using $k$-means.

Specifically, We denote the whole training set as $\mathcal{D}=\left\{(x_i, y_i)\right\}$.
%Images $\mathcal{X}=\left\{x_1, x_2, \cdots, x_n\right\}$ are extracted into feature vectors $\mathcal{V}$ through ViT-B pretrained on DINO~\cite{caron2021emerging}.
The feature $\mathbf{v}^i=\textit{f}(x_i)$ is extracted from each image $x_i$. %is -Base~\cite{dosovitskiy2020image} backbone pretrained with DINO~\cite{caron2021emerging}.
%Since DINO~\cite{caron2021emerging} is a self-supervised learning framework, the features $\mathcal{V}$ must be class-agnostic, which 
%is important for further class-specific task.
The $v_i$ extracted by DINO~\cite{caron2021emerging} is incapable of distinguishing between the fine-grained classes since there is no supervision during training, but it will provide a rough description of the image so that similar images will be clustered together. 
%The self-supervised vision transformer possesses the properties of an excellent nearest neighbors classifier and strong scene layout information, which is directly accessible to the last block of the model.
Then, the whole $\mathcal{D}$ is clustered into $K$ sub-datasets $\{\mathcal{D}^1, \mathcal{D}^2, \cdots, \mathcal{D}^K\}$ using $k$-means, each containing similar images and will be used for fine-grained category discovery later.

\subsection{Learning discriminative representations}
Since the images within each of the partitioned sub-dataset only have fine-grained differences with each other, and each sub-dataset naturally has different global statistics overall, we use a set of projectors $h_j(\cdot), j=1,\cdots,K$ to project features to each corresponding sub-spaces in which contrastive learning  will be performed.
Each projector can be considered an expert projector dedicated to learning fine-grained discriminative features from each sub-dataset.
Similar to Vaze~\etal~\cite{vaze2022generalized}, we apply both supervised contrastive loss and self-supervised contrastive loss to fine-tune the model.
Specifically, our proposed fine-grained self-supervised contrastive loss is
\begin{equation}\label{eq:fg_unsupcon}
    \mathcal{L}_{\text{fine}}^{u} = -  \sum_{k=1}^{K} \frac{1}{\left|\mathcal{B}^k\right|} \sum_{i\in \mathcal{B}^k} \log \frac{\exp(h_k(\mathbf{v}_i)\cdot h_k(\hat{\mathbf{v}}_i) / \tau) }{\sum_{j} \mathds{1}_{[j\ne i]} \exp(h_k(\mathbf{v}_i) \cdot h_k(\mathbf{v}_j) / \tau)}
\end{equation}
where $\mathcal{B}^k$ is a minibatch of images sampled from a partitioned dataset $\mathcal{D}^k$, $\mathbf{v}_i$ and $\hat{\mathbf{v}}_i$ are two views of one same image through data augmentation, and $\tau$ is the temperature parameter.

The fine-grained supervised contrastive loss is defined similarly
\begin{equation}\label{eq:fg_supcon}
    \mathcal{L}_{\text{fine}}^{l} = - \sum_{k=1}^{K} \frac{1}{\left|\mathcal{B}^k\right|} \sum_{i\in \mathcal{B}^k} \frac{1}{\left|\mathcal{N}(i)\right|} \sum_{q \in \mathcal{N}(i)} \log \frac{\exp(h_k(\mathbf{v}_i)\cdot h_k(\mathbf{v}_q) / \tau) }{\sum_j \mathds{1}_{[j \ne i]} \exp (h_k(\mathbf{v}_i) \cdot h_k({\mathbf{v}_j}) / \tau)}
\end{equation}
where $\mathcal{N}(i)$ is the set of indices for images with the same label as the anchor image $i$.

Thus, the overall loss we propose to learn fine-grained features is the combination of two losses defined above
\begin{equation}\label{eq:fg_overall}
    \mathcal{L}_{\text{fine}} = (1-\lambda) \mathcal{L}_{\text{fine}}^{u} + \lambda \mathcal{L}_{\text{fine}}^{l}
\end{equation}

%For our $K+1$ groups of data, we optimize their coarse-grained loss and fine-grained loss weighted by the hyperparameter $\alpha$. The overall loss is:
Together with the loss from Vaze~\etal~\cite{vaze2022generalized} defined in ~\cref{eq:total}, which can be viewed as a coarse-grained loss $\mathcal{L}_{\text{coarse}}$ compared to our proposed $\mathcal{L}_{\text{fine}}$, our optimization objective is
\begin{equation}\label{eq:overall}
\mathcal{L} = \mathcal{L}_{\text{coarse}} + \alpha \mathcal{L}_{\text{fine}}
\end{equation}
where $\alpha$ is a parameter to balance between our proposed $\mathcal{L}_{\text{fine}}$ and the original $\mathcal{L}_{\text{coarse}}$ from Vaze~\etal~\cite{vaze2022generalized}.
After the representation is learned, we run the semi-supervised $k$-means algorithm to obtain the cluster assignments of each sample.

\section{Experiments}
% datasets
\noindent \textbf{Datasets}. We evaluate our method on both generic image classification datasets and fine-grained datasets, with a special focus on the performance of the fine-grained image classification datasets. 
%including CIFAR-10~\cite{Krizhevsky09cifar}, CIFAR-100~\cite{Krizhevsky09cifar}, ImageNet-100~\cite{deng2009imagenet}. 
%For ImageNet-100, we randomly subsample 100 classes from ImageNet-1k.
%These are typical datasets to measure the performance on common classes.
Following previous works, we choose CIFAR-10/100~\cite{Krizhevsky09cifar}, ImageNet-100~\cite{deng2009imagenet} as the generic image classification datasets.
%Given that our method can model the distribution between coarse-grained and fine-grained representation, we further focus on the case that objects are similar in overall characteristics but are different in slight descriptions.
For fine-grained datasets we choose CUB-200~\cite{cub200}, Standford Cars~\cite{KrauseStarkDengFei-Fei_3DRR2013}, FGVC-Aircraft~\cite{maji13fine-grained}, and Oxford-IIIT Pet~\cite{parkhi2012cats}.
These fine-grained datasets contain categories from the same \textit{entry level} classes, \eg, birds, cars, aircrafts, and pets. These datasets can be more challenging for GCD methods requiring models to learn highly discriminative features~\cite{zhao2021novel}.
%These datasets better show the real-world of our life and include images that are closely related entities, \eg birds, pets and aircraft.
%Under this condition, the two-level representations can be more discriminative rule for the fine-grained GCD problem.
We split the training data into a labeled dataset and an unlabeled dataset by first dividing all classes equally into a seen class set and an unseen one then sampling 50\% images from the seen classes as unlabeled data so that the unlabeled set $\mathcal{D}^u$ contains images from both seen classes and unseen classes, while the labeled set only contains seen classes, the splits are presented in~\cref{tab:datasets}.%, while the labeled set $\mathcal{D}^l$ contain labeled images as shown in ~\cref{tab:datasets}.

\begin{table}[htb]
\footnotesize
\centering
\caption{Our dataset splits in the experiments.}
\label{tab:datasets}
\resizebox{\linewidth}{!}{
\begin{tabular}{ccccccccc}
\toprule
\multicolumn{2}{c}{\text{Dataset}}    & CIFAR10 & CIFAR100  & ImageNet-100  & CUB-200 & SCars & Aircraft & Pet  \\
\midrule
\multirow{2}{*}{Labelled}   & Classes & 5       &  80      & 50           & 100     & 98    & 50        & 19   \\
& Images  & 12.5k   & 20k & 31.9k  & 1498    & 2000  & 1666      & 942  \\
\hline
\multirow{2}{*}{Unlabelled} & Classes & 10      &  100       & 100     & 200     & 196   & 100       & 37  \\
& Images  & 37.5k   & 30k & 95.3k  & 4496    & 6144  & 5001      & 2738 \\
\bottomrule
\end{tabular}
}
\end{table}

% evaluation metric
\noindent \textbf{Evaluation metric}.
We employ the clustering accuracy ($ACC$) on the unlabeled set to measure the performance. 
The evaluation metric is defined as below
\begin{equation}\label{eq:acc}
ACC = \mathop{\max}_{p\in{\mathcal{P}(y^u)}} \frac{1}{N} \sum_{i=1}^{N} \mathbbm{1} \left\{y_i= p(\hat{y_i})\right\}
\end{equation}
where $\mathcal{P}$ is the set of all permutations that can match the clustering prediction $\hat{y_i}$ with the ground-truth label $y_i$, we use the Hungarian algorithm~\cite{kuhn1955hungarian} to find the best permutation, and $N$ is the number of images in the unlabeled set.
Following ~\cite{vaze2022generalized}, we use the metric on three different sets, including `All' referring to the entire unlabeled set $\mathcal{D}^u$, `Old' referring to instances in $\mathcal{D}^u$ belonging to classes in $\mathcal{C}^l$ and `New' referring to instances in $\mathcal{D}^u$ belonging to $\mathcal{C}^u \setminus \mathcal{C}^l$.

\subsection{Implementation details}
We follow the implementation of ~\cite{vaze2022generalized} to use ViT-B-16~\cite{dosovitskiy2020image} as the backbone of our method.
We initialize the model with the parameters pretrained by DINO~\cite{caron2021emerging} on ImageNet and only fine-tune the final transformer block while other blocks are frozen.
%Our projection heads are the same as ~\cite{caron2021emerging} and will be discarded when testing. 
%We duplicate the projection heads for the number of $K+1$, and initialize them separately.
%We first cluster the features that pass through our model using $k$-means and split the entire traning dataset into several sub-groups.
%We then train our model using the sub-groups and the entire training datasets together.
%The number of sub-groups K is set to 8. 
We implement the projection heads as three layer MLPs following DINO~\cite{caron2021emerging}, these projection heads will be discarded when testing.
The batch size for the entire training dataset is set to 256 and the batch size of all the sub-datasets is set to 32.
For the ImageNet dataset, all models are trained for 60 epochs while for other datasets, models are trained for 200 epochs. 
We set $\alpha$ to be $0.1$ by default.
Similar to ~\cite{vaze2022generalized}, we use a base learning rate of $0.1$ with a cosine annealing schedule and set $\lambda$ to 0.35. 
For a fair comparison with existing methods, we use the same semi-supervised $k$-means method as ~\cite{vaze2022generalized} to do the evaluation.

\subsection{Comparison with the State-of-the-Art}
%We compare our method with the previous state-of-the-art on three generic image classification datasets reported in ~\cref{tab:generic} and four fine-grained image classification datasets reported in ~\cref{tab:fine_grained}.
We first compare \OURS~with the state-of-the-art methods on both generic image classification benchmarks and fine-grained image classification benchmarks.
The $k$-means method in the tables refers to running $k$-means directly on the features extracted from DINO without any further finetuning.
RankStats+ and UNO+ are two methods modified from two competitive baselines for NCD and adopted to the GCD setting, \ie RankStats~\cite{han2021autonovel} and UNO~\cite{fini2021unified}.

The results on generic image classification benchmarks are shown in~\cref{tab:generic}.
On all the datasets we tested, \OURS~shows the best performance on `All', showing that our method could improve upon previous works.
\OURS~also achieves comparable results with other methods on the other subsets as `Old' and `New'.
It should be noticed the best performance on ImageNet-100 `New' subset is achieved by naively running a $k$-means on DINO features, suggesting that the original features can already represent the unlabeled categories well, and \OURS~achieves the closest performance compared to this baseline, showing that unlike existing method potentially introducing damage to original feature quality which results in significant performance drop, our method can best preserve the high quality of original features.

% CIFAR10
\begin{table}[htb]
\small
\centering
\caption{Results on generic datasets.% \OURS~yields the highest performance on most generic benchmarks
}
\label{tab:generic}
% \resizebox{0.33\linewidth}{!}{ %< auto-adjusts font size to fill line
\begin{tabular}{lccccccccc}
\toprule
& \multicolumn{3}{c}{CIFAR10} & \multicolumn{3}{c}{CIFAR100} & \multicolumn{3}{c}{ImageNet-100} \\
\cmidrule(rl){2-4}
\cmidrule(rl){5-7}
\cmidrule(rl){8-10}
Method       & All      & Old     & New        & All       & Old       & New       & All       & Old       & New  \\
\midrule
$k$-means~\cite{MackQueen67_Kmeans}
& 83.6
& 85.7
& 82.5
& 52.0
& 52.2
& 50.8
& 72.7
& 75.5
& \textbf{71.3}
\\
RankStats+
&  46.8
&  19.2
&  60.5
&  58.2
&  77.6
&  19.3
&  37.1
&  61.6
&  24.8
\\
UNO+ 
&  68.6
&  \textbf{98.3}
&  53.8
&  69.5
&  80.6
&  47.2
&  70.3
&  \textbf{95.0}
&  57.9
\\
GCD~\cite{vaze2022generalized}
& 91.5
& 97.9
& 88.2
& 73.0
& 76.2
& \textbf{66.5}
& 74.1
& 89.8
& 66.3
\\
\midrule
\OURS
& \textbf{96.0}
& 97.3
& \textbf{95.4}
& \textbf{74.2}
& \textbf{81.2}
& 60.3
& \textbf{77.6}
& 93.5
& 69.7
\\
\bottomrule
\end{tabular}
% }
\end{table}

% fine-grained results.
%\TODO{pet k-means}

\begin{table}[htb]
\footnotesize
\centering
\caption{Results on fine-grained datasets.% \OURS~out-performs existing methods by a large margin on most fine-grained benchmarks
}
\label{tab:fine_grained}
\resizebox{\linewidth}{!}{ %< auto-adjusts font size to fill line
\begin{tabular}{lcccccccccccc}
\toprule
 & \multicolumn{3}{c}{CUB-200} & \multicolumn{3}{c}{Stanford-Cars} & \multicolumn{3}{c}{FGVC-Aircraft} & \multicolumn{3}{c}{Oxford-Pet} \\
\cmidrule(rl){2-4}
\cmidrule(rl){5-7}
\cmidrule(rl){8-10}
\cmidrule(rl){11-13}
Method       & All      & Old     & New        & All       & Old       & New       & All       & Old       & New   & All        & Old       & New \\
\midrule
$k$-means~\cite{MackQueen67_Kmeans}
& 34.3
& 38.9
& 32.1
& 12.8
& 10.6
& 13.8
& 16.0
& 14.4
& 16.8
& 77.1
& 70.1
& 80.7
\\
RankStats+
& 33.3
& 51.6
& 24.2
& 28.3
& 61.8
& 12.1
& 26.9
& 36.4
& 22.2
& $\text{-}$
& $\text{-}$
& $\text{-}$
\\
UNO+ 
& 35.1
& 49.0
& 28.1
& 35.5
& \textbf{70.5}
& 18.6
& 40.3
& \textbf{56.4}
& 32.2
& $\text{-}$
& $\text{-}$
& $\text{-}$
\\
GCD~\cite{vaze2022generalized}
& 51.3
& \textbf{56.6}
& 48.7
& 39.0
& 57.6
& 29.9
& 45.0
& 41.1
& 46.9
& 80.2
& 85.1
& 77.6
\\
\midrule
\OURS
& \textbf{52.1}
& 54.3
& \textbf{51.0}
& \textbf{40.5}
& 58.8
& \textbf{31.7}
& \textbf{47.7}
& 44.4
& \textbf{49.4}
& \textbf{86.7}
& \textbf{91.5}
& \textbf{84.1}
\\
\bottomrule
\end{tabular}
}
\end{table}

We present the results on fine-grained image classification benchmarks in~\cref{tab:fine_grained}.
Our method shows the best performance on the `All' and `New' with all four datasets we tested while achieving comparable results on `Old', indicating the effectiveness of our method for fine-grained category discovery.

%\subsection{Bias-Variance analysis}

\subsection{Ablation study}
We perform the ablation study by adjusting each element of our method to inspect the effectiveness of them. 
For quicker evaluation, we use two fine-grained datasets, \ie CUB-200 and Standford Cars, and train the model for 100 epochs to ablate the performance.

\paragraph{Fine-grained and coarse-grained loss.}
%~\cref{tab:ablate_cg} shows the absence of coarse-grained loss will hurt performance. 
\cref{tab:ablate_cg} presents the performance of using different combinations of loss terms.
We observed that with additional supervision from the coarse-grained loss, the $ACC$ is improved by $3.3\text{-}4.0\%$ on CUB-200 and $15.4\text{-}28.5\%$ on Standford Cars.
As combining the fine-grained and coarse-grained losses achieves the best performance, it is proved that our proposed method to learn fine-grained features improves GCD methods' performance in fine-grained benchmarks.
%Our method proposes the data splitting to use a combination of coarse-grained set and fine-grained set altogether when training. 
%The result demonstrates that only with the fine-grained loss, the model can just learn the knowledge inside each sub-group, which is too discriminative for the model to find the relationship among sub-groups. 

\begin{table}[htb]
\small
\centering
\caption{Ablation study of fine-grained loss and coarse-grained loss. 
%\TODOTAB{only with $L_{cg}$ for 100 epochs.}
}
\label{tab:ablate_cg}
%\resizebox{\linewidth}{!}{
\begin{tabular}{cccccccc}
\toprule
\multirow{2}{*}{$\mathcal{L}_{\text{fine}}$}   & \multirow{2}{*}{$\mathcal{L}_{\text{coarse}}$}  &   \multicolumn{3}{c}{CUB-200}  & \multicolumn{3}{c}{Stanford-Cars}  \\
\cmidrule(rl){3-5}
\cmidrule(rl){6-8}
                                     &                        
              &   All      & Old     & New        & All       & Old       & New     \\
\midrule
\checkmark 
& 
& 48.0
& 50.5
& 46.8
& 21.3
& 30.6
& 16.8 
\\
 
& \checkmark
& 49.9
& 53.4
& 48.2
& 37.1
& 57.9
& 27.0 
\\
\checkmark
& \checkmark
& \textbf{51.8}
& \textbf{53.8}
& \textbf{50.8}
& \textbf{41.0}
& \textbf{59.1}
& \textbf{32.2}
\\
\bottomrule
\end{tabular}
%}
\end{table}

\paragraph{The weight of fine-grained loss.}
We analyze the choice of the weight $\alpha$ for fine-grained loss in~\cref{tab:ablate_weight}. 
We find that \OURS~can consistently outperform the baseline($\alpha = 0$) with different $\alpha$, showing the robust effectiveness of our method.
The best result is achieved with $\alpha=0.4$ on CUB-200 and with $\alpha=0.2$ on Standford Cars.

\begin{table}[htb]
\small
\centering
\caption{Ablation study on the weight $\alpha$ of loss. $\alpha = 0$ is the baseline(Vaze~\etal~\cite{vaze2022generalized}).}
\label{tab:ablate_weight}
%\resizebox{\linewidth}{!}{
\begin{tabular}{ccccccc}
\toprule
\multirow{2}{*}{$\alpha$}    &   \multicolumn{3}{c}{CUB-200}  & \multicolumn{3}{c}{Stanford-Cars}  \\
\cmidrule(rl){2-4}
\cmidrule(rl){5-7}
   &             All      & Old     & New        & All       &            Old       & New     \\
\midrule
0
& 49.9
& 53.4
& 48.2
& 37.1
& 57.9
& 27.0  
\\
0.1 
& 51.8
& 53.8
& 50.8
& 41.0
& 59.1
& 32.2
\\
0.2
& 51.6
& 54.5
& 50.2
& \textbf{42.4}
& \textbf{63.0}
& \textbf{32.4}
\\
0.4
& \textbf{53.4}
& \textbf{58.6}
& \textbf{50.9}
& 41.1
& 61.2
& 31.4
% \\
% 0.8
% & 51.9
% & \textbf{58.8}
% & 48.4
% & 36.4
% & 57.5
% & 26.2
\\
\bottomrule
\end{tabular}
%}
\end{table}

\begin{figure*}[t]
\begin{center}
\includegraphics[width=1.\linewidth]{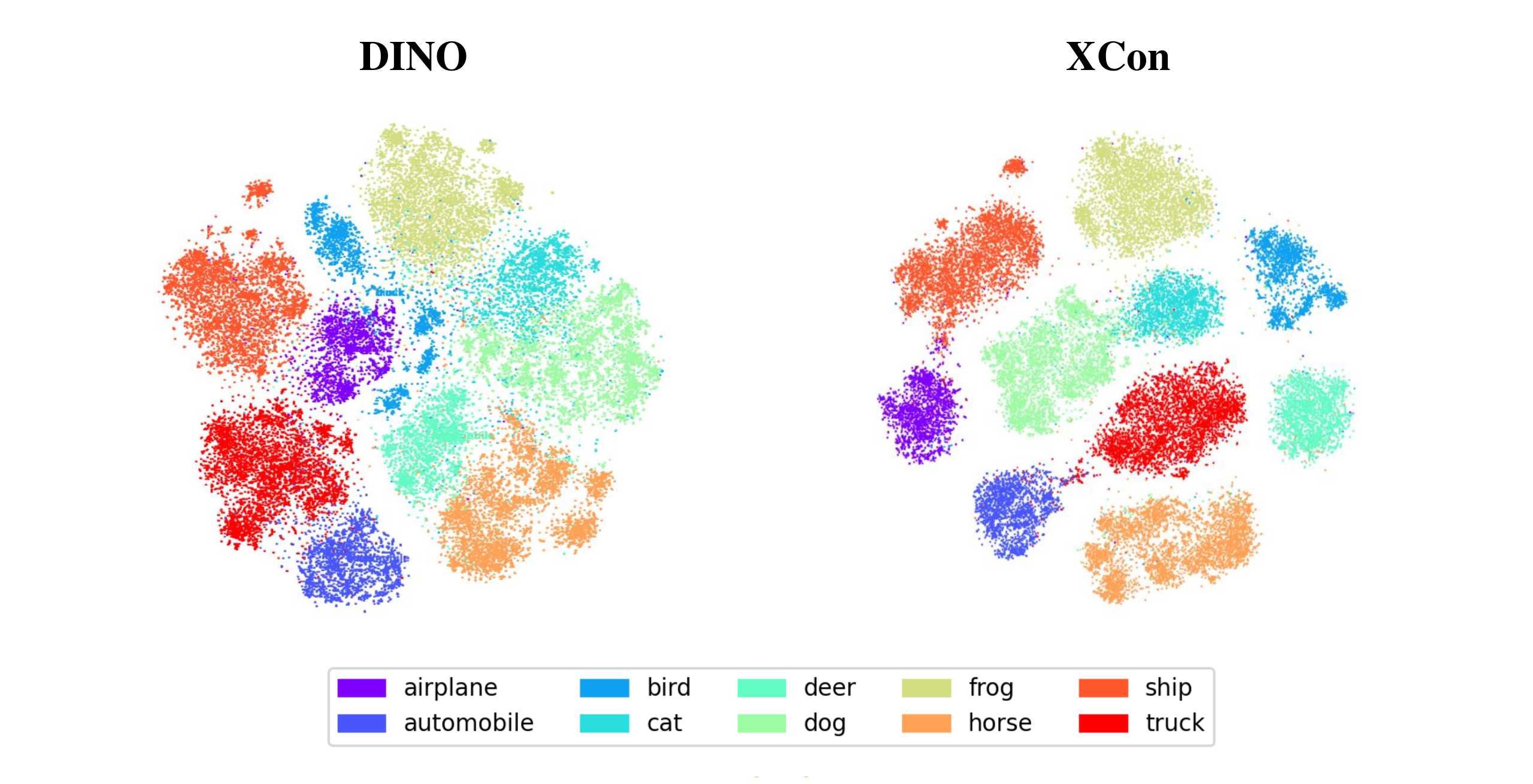}
\end{center}
\vspace{-22pt}
\caption{Feature visualization on CIFAR10 with TSNE. }
\label{fig:tsne}
\end{figure*}

\paragraph{The number of sub-datasets.} The effect of the sub-dataset number is illustrated in ~\cref{tab:ablate_k}.
%The fact is that $ACC$ is increased as long as $K>1$, and the best choice of $K$ is varied by datasets.
Although the performance of \OURS~is consistently better than the baseline's, it still varies greatly depending on the number of sub-datasets.
When $K=2$, it can reach the highest on the `Old' set, but the lowest on the `New' set, that means with two groups, the overall difference between features is not so great inside each group that the model tends to focus more on the existing coarse-grained knowledge learned from the seen classes.

\begin{table}[ht]
\small
\centering
\caption{Ablation study on the number $K$ of split sub-groups.}
\label{tab:ablate_k}
%\resizebox{\linewidth}{!}{
\begin{tabular}{ccccccc}
\toprule
\multirow{2}{*}{$K$}    &   \multicolumn{3}{c}{CUB-200}  & \multicolumn{3}{c}{Stanford-Cars}  \\
\cmidrule(rl){2-4}
\cmidrule(rl){5-7}
   &             All      & Old     & New        & All       &            Old       & New     \\
\midrule
1 
& 49.9
& 53.4
& 48.2
& 37.1
& 57.9
& 27.0  
\\
2
& 51.4
& \textbf{59.3}
& 47.4
& 40.9
& \textbf{61.0}
& 31.1
\\
4
& 51.7
& 54.6
& 50.2
& 39.8
& 55.3
& 32.3
\\
6
& 50.3
& 51.9
& 49.5
& \textbf{42.1}
& 60.7
& \textbf{33.1}
\\
8
& \textbf{51.8}
& 53.8
& \textbf{50.8}
& 41.0
& 59.1
& 32.2
\\
\bottomrule
\end{tabular}
%}
\end{table}

We further visualize the feature spaces with TSNE~\cite{van2014accelerating} on CIFAR10 by mapping the features into two dimensions for a more direct qualitative analysis.
In ~\cref{fig:tsne}, we cluster the unlabeled data and compare results from the initial model(DINO) with ones from our method(\OURS).
It is clear that the improvement from DINO to our method is significant.
DINO can cluster the features into $10$ groups roughly, but many samples appear in groups that correspond to other classes.
In contrast to DINO, with our model, we can see clear boundaries between different groups, and each group is corresponding to one certain category in CIFAR10.

\section{Conclusion}
In this paper, we propose \OURS~ to address the problem of generalized category discovery with fine-grained image classification benchmarks.
\OURS~ first partitions the dataset into $K$ sub-dataset using $k$-means clustering on a self-supervised representation.
Each partitioned sub-dataset can be seen as a subset of images that are visually similar and have close coarse-grained representation so that contrastive learning within each of these sub-datasets will force the model to learn fine-grained discriminative features that can help discover fine-grained categories.
Experiments on four fine-grained image classification benchmarks show clear performance improvements of \OURS, validating the effectiveness of our method.

\section*{Acknowledge}
The author would like to acknowledge compute support from LunarAI.

\bibliography{bmvcbib}

\clearpage
\appendix

% different self-supervised representation.
% Use estimated class numbers.
% estimating class numbers using our representations.

% loss ablation: 
% unsupervised loss only.
% unsupervised loss + supervised loss.

\section{Different self-supervised representations}
The self-supervised representation provides us the information to partition the training data into $k$ expert sub-datasets, so we analyze the performance of our method by fine-tuning different pretrained representations of other self-supervised ViT models, \ie MoCo v3~\cite{chen2021empirical} and MAE~\cite{he2022masked}.
We initialize the ViT-B-16 model~\cite{dosovitskiy2020image} with the parameters pretrained on ImageNet-1k by MoCo v3 for 300 epochs and by MAE for 800 epochs respectively.
The result in \cref{tab:repre} shows that with the self-supervised representation of DINO~\cite{caron2021emerging}, our method performs $7.3-20.7\%$ better than the other two on CUB-200 and $1.8-22.4\%$ better on Standford-Cars.
We observe that DINO still shows the best performance on clustering the data based on class-irrelevant informations.
%Surprisingly, \OURS with MAE can outperform the GCD method~\cite{vaze2022generalized} under the same setting of training for 100 epochs.

\begin{table}[htb]
\small
\centering
\caption{Results with different self-supervised representations.}
\label{tab:repre}
%\resizebox{\linewidth}{!}{
\begin{tabular}{ccccccc}
\toprule
\multirow{2}{*}{\makecell[c]{self-supervised \\ ViT model}}    &   \multicolumn{3}{c}{CUB-200}  & \multicolumn{3}{c}{Stanford-Cars}  \\
\cmidrule(rl){2-4}
\cmidrule(rl){5-7}
   &             All      & Old     & New        & All       &            Old       & New     \\
\midrule
DINO
& \textbf{51.8}
& \textbf{53.8}
& \textbf{50.8}
& \textbf{41.0}
& \textbf{59.1}
& \textbf{32.2}  
\\
MoCo v3
& 37.6
& 42.8
& 35.1
& 24.7
& 36.7
& 18.9
\\
MAE
& 35.5
& 46.5
& 30.1
& 38.7
& 56.0
& 30.4
\\
\bottomrule
\end{tabular}
%}
\end{table}

\section{Estimating the number of classes}
As a more realistic scenario, the prior knowledge of the number of classes is unknown in the GCD.
We follow the method in ~\cite{vaze2022generalized} to estimate the number of classes in the unlabeled dataset by leveraging the information of the labeled dataset.
We compare our estimated number of classes in unlabeled data $\left|\mathcal{\hat{C}}^u\right|$ with the ground truth number of classes in unlabeled data $\left|\mathcal{C}^u\right|$ in \cref{tab:est_k}.
We find that on Standfor-Cars and FGVC-Aircraft, the number of classes estimated by our method is significantly closer to the ground truth compared with GCD~\cite{vaze2022generalized}.
Our method tends to show better performance on fine-grained datasets, given that the dataset partitioning can help the model learn more discriminative features when facing the more challenging datasets that have little obvious difference.

\begin{table}[htb]
\footnotesize
\centering
\caption{
Estimation of the number of classes in unlabeled data.
}
\label{tab:est_k}
\resizebox{\linewidth}{!}{
\begin{tabular}{lcccccccc}
\toprule
   &CIFAR10 &CIFAR100 &ImageNet-100 & CUB-200 & Standford-Cars & FGVC-Aircraft  & Oxford-Pet \\
\midrule
Ground truth  
&10
&100
&100
&200
&196
&100
&37
\\
GCD~\cite{vaze2022generalized}
&9
&100
&109
&231
&230
&80
&34
\\
\OURS
&8
&97
&109
&236
&206
&101
&34
\\
\bottomrule
\end{tabular}
}

\end{table}

\section{Performance with estimated class number}
We use the class number estimated in \cref{tab:est_k} to evaluate our method, displaying the performance of our method when the unlabeled class number is unavailable.
We report the results on generic image classification benchmarks in \cref{tab:est_generic} and the results on fine-grained image classification benchmarks in \cref{tab:est_fg}. 
With our estimated class number $\left|\mathcal{\hat{C}}^u\right|$, our method performs better on Standford-Cars and also reaches comparable results on the other five datasets except CIFAR10, which shows that our method is also promising under the more realistic condition.
\begin{table}[htb]
\footnotesize
\centering
\caption{Results on generic datasets with our estimated class number.
}
\label{tab:est_generic}
%\resizebox{\linewidth}{!}{ %< auto-adjusts font size to fill line
\begin{tabular}{cccccccccc}
\toprule
\multirow{2}{*}{known $C^u$} & \multicolumn{3}{c}{CIFAR10} & \multicolumn{3}{c}{CIFAR100} & \multicolumn{3}{c}{ImageNet-100} \\
\cmidrule(rl){2-4}
\cmidrule(rl){5-7}
\cmidrule(rl){8-10}
      & All      & Old     & New        & All       & Old       & New       & All       & Old       & New  \\
\midrule
\cmark
& \textbf{96.0}
& 97.3
& \textbf{95.4}
& \textbf{74.2}
& \textbf{81.2}
& \textbf{60.3}
& \textbf{77.6}
& \textbf{93.5}
& \textbf{69.7}
\\
\xmark
& 70.1
& \textbf{97.4}
& 56.5
& 72.5
& 80.3
& 56.8
& 75.6
& 91.5
& 67.6
\\
\bottomrule
\end{tabular}
%}
\end{table}

\begin{table}[htb]
\footnotesize
\centering
\caption{Results on fine-grained datasets with our estimated class number.
}
\label{tab:est_fg}
\resizebox{\linewidth}{!}{ %< auto-adjusts font size to fill line
\begin{tabular}{ccccccccccccc}
\toprule
\multirow{2}{*}{known $C^u$}  & \multicolumn{3}{c}{CUB-200} & \multicolumn{3}{c}{Stanford-Cars} & \multicolumn{3}{c}{FGVC-Aircraft} & \multicolumn{3}{c}{Oxford-Pet} \\
\cmidrule(rl){2-4}
\cmidrule(rl){5-7}
\cmidrule(rl){8-10}
\cmidrule(rl){11-13}
      & All      & Old     & New        & All       & Old       & New       & All       & Old       & New   & All        & Old       & New \\
\midrule
\cmark
& \textbf{52.1}
& 54.3
& \textbf{51.0}
& 40.5
& 58.8
& 31.7
& \textbf{47.7}
& 44.4
& \textbf{49.4}
& \textbf{86.7}
& \textbf{91.5}
& \textbf{84.1}
\\
\xmark
& 51.0
& \textbf{57.8}
& 47.6
& \textbf{41.3}
& \textbf{58.8}
& \textbf{32.8}
& 46.1
& \textbf{47.6}
& 45.3
& 82.1
& 81.7
& 82.4
\\
\bottomrule
\end{tabular}
}
\end{table}

\section{Ablation on contrastive fine-tuning}
We further ablate the components of contrastive loss in \cref{tab:ablate_con}.
We find that only with unsupervised contrastive loss, \ie $\lambda=0$, the $ACC$ drops $21.5-23.6\%$ on CUB-200 and $22.2-46.6\%$ on Standford-Cars, which means the combination of supervised contrastive loss and unsupervised contrastive loss with the balanced parmeter $\lambda=0.35$ is necessary and can reach the best performance.

\begin{table}[htb]
\small
\centering
\caption{Ablation study of contrastive loss.
}
\label{tab:ablate_con}
%\resizebox{\linewidth}{!}{
\begin{tabular}{ccccccc}
\toprule
\multirow{2}{*}{$\lambda$}    &   \multicolumn{3}{c}{CUB-200}  & \multicolumn{3}{c}{Stanford-Cars}  \\
\cmidrule(rl){2-4}
\cmidrule(rl){5-7}  
              &   All      & Old     & New        & All       & Old       & New     \\
\midrule
0 
& 29.6
& 30.2
& 29.3
& 10.8
& 12.5
& 10.0
\\
0.35
& \textbf{51.8}
& \textbf{53.8}
& \textbf{50.8}
& \textbf{41.0}
& \textbf{59.1}
& \textbf{32.2}
\\
\bottomrule
\end{tabular}
%}
\end{table}

% \bibliography{bmvcbib}

\end{document}